\documentclass[runningheads]{llncs}
\usepackage[T1]{fontenc}
\usepackage{graphicx}
\usepackage{booktabs}
\usepackage[misc]{ifsym}
\newcommand{\corr}{(\Letter)}
\usepackage{mwe}

\usepackage{orcidlink}
\usepackage[utf8]{inputenc} % allow utf-8 input
\usepackage[T1]{fontenc}    % use 8-bit T1 fonts
\usepackage{hyperref}       % hyperlinks
\usepackage{url}            % simple URL typesetting
\usepackage{booktabs}       % professional-quality tables
\usepackage{amsfonts}       % blackboard math symbols
\usepackage{nicefrac}       % compact symbols for 1/2, etc.
\usepackage{microtype}      % microtypography
\usepackage{xcolor}         % colors
\usepackage[ruled,vlined,linesnumbered,commentsnumbered]{algorithm2e}

\usepackage{microtype}
\usepackage{graphicx}
\usepackage{multirow}
\usepackage{subfig}
\usepackage{booktabs} % for professional tables
\usepackage{amsmath}
\usepackage{amssymb}
\usepackage{mathtools}

\usepackage{amsthm}
\usepackage{enumitem}
\usepackage[capitalize,noabbrev]{cleveref}
\usepackage{verbatim}
\usepackage{comment}
\usepackage{stfloats}
\usepackage{tikz}

\definecolor{MyDarkBlue}{rgb}{0,0.5,1}
\definecolor{MyDarkGreen}{rgb}{0.02,0.6,0.02}
\definecolor{MyDarkRed}{rgb}{0.8,0.02,0.02}
\definecolor{MyDarkOrange}{rgb}{0.40,0.2,0.02}
\definecolor{MyPurple}{RGB}{111,0,255}
\definecolor{MyRed}{rgb}{1.0,0.0,0.0}
\definecolor{MyGold}{rgb}{0.75,0.6,0.12}
\definecolor{MyDarkgray}{rgb}{0.66, 0.66, 0.66}

\newcommand{\model}{Vid2Act}
\begin{document}

\title{Model-Based Reinforcement Learning with Multi-Task Offline Pretraining}

% \titlerunning{Underwater Basket Weaving Under Extreme Pressure}
% If the full title of your paper is short enough to also fit in the running head, you can omit the abbreviated paper title here. You can check as follows: if you comment out the \titlerunning line, something will appear in the header of all odd-numbered pages of your PDF from page 3 onward. This something is either the full title (in which case all is well), or the error message "Title Suppressed Due to Excessive Length". If this error message appears, you're going to want to provide an abbreviated title within the \titlerunning command, because if you won't do it, Springer will do it for you.

%N.B.: Author information (both in the \author{} and \authorrunning{} command) should only be present in the Camera-Ready Version of your paper. The version that you initially submit for review, ought to be double-blind. So, when initially submitting your paper, use:
% \author{Author information scrubbed for double-blind reviewing}
% \author{Andr\'e Lauren Benjamin\inst{1} \and
% Calvin Cordozar Broadus Jr.\inst{2,3} \corr \and
% Antwan Andr\'e Patton\inst{1}\orcidID{0000-1111-2222-3333}}

\newcommand*\samethanks[1][\value{footnote}]{\footnotemark[#1]}

\author{Minting Pan\thanks{Equal contribution.
 \\  Code available at \url{https://github.com/panmt/Vid2Act}} 
\and Yitao Zheng\samethanks   \and
Yunbo Wang \corr \and
Xiaokang Yang}

% \title{My title}

% \author{Minting Pan\footnotemark[1] \and 
%         Yitao Zheng\footnotemark[1]  \and
%         Yunbo Wang \and
%         Xiaokang Yang}
% \footnotetext[1]{Equal contribution. Code available at \url{https://github.com/panmt/Iso-Dream}}
% You may leave out the orcidID information, if you want to.
% Use \corr to indicate the corresponding author. Note the spacing around the \corr command. Only one author can be the corresponding author.

%N.B.: comment out the \authorrunning{} command for the double-blind version of your paper submitted for review. Later, if your paper is accepted, use the command for the Camera-Ready Version.
\authorrunning{M. Pan et al.}
% If there are two authors, write 'A.L. Benjamin and C.C. Broadus Jr.'
% If there are more than two authors, '[...] et al.' is used.

\institute{MoE Key Lab of Artificial Intelligence, AI Institute, Shanghai Jiao Tong University \email{\{panmt53, iorisou0826, yunbow, xkyang\}@sjtu.edu.cn}}

\tocauthor{Minting Pan, Yitao Zheng, Yunbo Wang, Xiaokang Yang}
\toctitle{Model-Based Reinforcement Learning with Multi-Task Offline Pretraining}

\maketitle              % typeset the header of the contribution

\begin{abstract}

Pretraining reinforcement learning (RL) models on offline datasets is a promising way to improve their training efficiency in online tasks, but challenging due to the inherent mismatch in dynamics and behaviors across various tasks. We present a model-based RL method that learns to transfer potentially useful dynamics and action demonstrations from offline data to a novel task. The main idea is to use the world models not only as simulators for behavior learning but also as tools to measure the task relevance for both dynamics representation transfer and policy transfer. We build a time-varying, domain-selective distillation loss to generate a set of offline-to-online similarity weights. These weights serve two purposes: (i) adaptively transferring the task-agnostic knowledge of physical dynamics to facilitate world model training, and (ii) learning to replay relevant source actions to guide the target policy. We demonstrate the advantages of our approach compared with the state-of-the-art methods in Meta-World and DeepMind Control Suite.

\end{abstract}

\section{Introduction}

Reinforcement learning (RL) approaches have made significant advancements in solving a wide range of sequential control problems~\cite{ebert2018visual,sekar2020planning,laskin2020reinforcement}. 
In the realm of visual RL, agents need to not only conduct representation learning from raw image inputs but also perform behavior learning in the learned state space, which requires a large number of interactions with an online environment and limits the applications in the real world.
Recently, model-based RL algorithms have greatly improved sample efficiency by concurrently learning a differentiable simulator of the environment (\textit{i.e.}, the world model), and using imagined rollouts generated by the world model for policy optimization~\cite{kaiser2019model,hafner2019dream}.
Nevertheless, the process of training an effective world model from scratch remains a time-consuming and challenging pursuit, often yielding less generalizable representations.

To address this problem, many recent approaches~\cite{stooke2021decoupling,sun2023smart,xu2022feasibility,taiga2022investigating} adopt the \textit{pretraining and finetuning} paradigm to pre-learn representation models on off-the-shelf offline datasets and transfer the learned prior knowledge to a novel online RL domain. 
For example, SMART~\cite{sun2023smart} exploits a Transformer model to learn generalizable visual representations from reward-free, offline interaction data under a control-centric pretraining objective.
Similarly, our focus lies in leveraging multi-task offline data without reward to improve the visual RL performance in a novel online task. 
However, it is crucial to recognize that, despite the effectiveness of the pretraining method, a straightforward finetuning method may still suffer from the potential discrepancy in visual observations, physical dynamics, or even action spaces across task domains.
Unlike SMART, our method aims to:
\begin{enumerate}[leftmargin=*]
    \item \textit{Adaptively identify the relevance between offline and online tasks in an unsupervised manner, allowing for positive domain transfer even when some offline data may seem unrelated.}
    \vspace{3pt} 
    \item \textit{Exploit relevant actions from the offline datasets to effectively guide and enhance the policy optimization process for the new task.}
\end{enumerate}

\begin{figure}[t]
\begin{center}
\centerline{\includegraphics[width=\columnwidth]{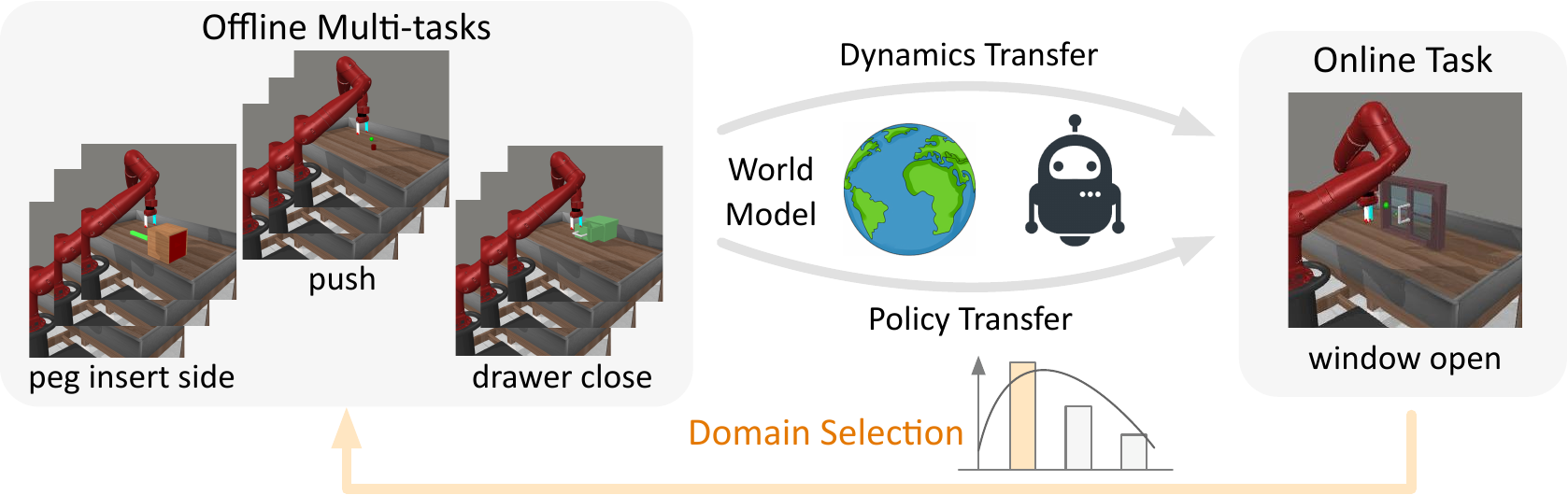}}
\caption{
We aim to build an offline-to-online transfer RL agent for visual control problems, which is challenging due to the discrepancies between the target task and the source tasks from which the offline datasets are collected. 
The key idea of our approach is to leverage the world models to enable positive knowledge transfer through \textit{domain-selective dynamics distillation} and \textit{behavior guidance}.
}
\label{fig:intro}
\end{center}
\vspace{-20pt}
\end{figure}

As shown in Fig. \ref{fig:intro}, we propose a new \textit{domain-selective} transfer RL approach called \model{} to reduce the potential discrepancies between the pretraining stage and the transferring stage.
In the pretraining stage, we exploit reward-free offline trajectories of image-action pairs to train a mixture world model, which learns the task-specific observation-to-state mapping functions and state-to-state transition functions based on task index for different source tasks.
In the transferring stage, instead of performing direct finetuning, we leverage the mixture world model as the teacher model to provide flexible regularization to the representation learning process of the target-domain agent.
This is achieved through a domain-selective distillation loss, where we learn a set of importance weights over the teacher model with different label indexes to adaptively transfer the prior knowledge of physical dynamics gathered from the offline data to the target world model.

In addition to their impact on representation learning, the importance weights also directly contribute to the policy optimization process conducted over the imaginations of the target world model.
Specifically, \model{} incorporates a ``\textit{generative action replay}'' module. During behavior learning, it serves to reproduce source-domain actions based on the target-domain states, which have been aligned to the corresponding source-domain state spaces by the distillation loss.
By reusing the importance weights, we can dynamically select the most relevant source task at different time steps and replay its source expert behaviors to provide effective guidance for target policy improvement.

In summary, the main technical contributions of \model{} are as follows:
\begin{itemize}[leftmargin=*]
    \item Our work introduces a novel pretraining and finetuning pipeline for visual model-based RL. It transfers the dynamics from multiple source tasks with a set of importance weights learned by the world model.
    \vspace{3pt}
    \item \model{} presents a novel domain-selective behavior learning scheme that identifies potentially valuable source actions and employs them as exemplar guidance for the target policy.
\end{itemize}

We evaluate \model{} on the Meta-World benchmark~\cite{yu2020meta} and the DeepMind Control Suite~\cite{tassa2018deepmind}. 
Our approach shows remarkable performance improvements over both the vanilla model-based RL baselines like DreamerV2~\cite{hafner2020mastering} and existing unsupervised pretraining methods for transfer RL, such as APV~\cite{seo2022reinforcement} and SMART~\cite{sun2023smart}. 
Importantly, our experimental results consistently demonstrate that \model{} can achieve positive domain transfer, even when the available source offline data seems less relevant to the target task.

\section{Related Work}

\noindent \textbf{Visual RL.}
In visual control tasks, the agent needs to learn policy from high-dimensional and complex observations.
Learning generalized representation by either unsupervised~\cite{gelada2019deepmdp,srinivas2020curl,stooke2021decoupling,yarats2021improving} or self-supervised manners~\cite{choudhary2022spatial,yang2022self,ze2023visual}, is a natural way to learn an auxiliary encoder of images for visual control tasks. Prior approaches consist of model-based methods to optimize latent dynamics model~\cite{hafner2019dream,hafner2019learning,hafner2020mastering,pan2022iso}, and model-free methods to utilize data augmentation~\cite{laskin2020reinforcement,chos2p} and contrastive representation learning~\cite{anand2019unsupervised,srinivas2020curl,nair2022r3m,li2023contrastive}. Similar to our work, several methods pretrain RL models on offline datasets and then finetune them on the online target task~\cite{dwibedi2018learning,srinivas2020curl,stooke2021decoupling,schwarzer2021pretraining,seo2022reinforcement}. Except for not bridging the domain gap between pretraining source data and RL tasks, they have shown attractive performance on vision-based RL tasks. In our framework, we do not directly finetune the parameters of the pretrained models, but rather learn more useful world models by distillation technique.  

\vspace{5pt}
\noindent \textbf{Transfer RL.}
Previous experiences across a diverse range of tasks can be beneficial in solving online control tasks, even when encountering them for the first time. 
% long2015learning,gupta2016cross,rebuffi2017learning,
To quickly leverage the past information to the new environments, many transfer learning approaches~\cite{hester2018deep,liu2019knowledge,yao2020unsupervised,yang2021representation,kadokawa2023cyclic} are proposed to bridge the gap across different tasks or domains. 
APV~\cite{seo2022reinforcement} employs action-free videos of multiple domains to pretrain an action-free recurrent state-space model (RSSM), which focuses on learning visual representation from offline datasets.
XTRA~\cite{xu2022feasibility} proposes a framework based on EfficientZero~\cite{2021Mastering} to use multiple offline tasks with rewards both in pretraining and finetuning stages for cross-task transfer.
Recently, some methods leveraging Transformer have been proposed to facilitate transfer learning in control tasks~\cite{xie2023future,sun2023smart}. SMART~\cite{sun2023smart} designs a control-centric pretraining objective for Decision Transformers~\cite{chen2021decision} to capture the common essential information relevant to short-term control and long-term control across tasks. 
A work closely related to our approach is Knowledge Flow~\cite{liu2019knowledge}, which involves training multiple teacher models and distilling knowledge from their layers to a student model.
In our work, we propose a domain-selective distillation strategy to fully utilize both the dynamics and action information from the source tasks. It introduces a more flexible way to adaptively transfer useful knowledge to help downstream tasks.

\section{Problem Formulation}

In the visual control task, the agent learns the behavior policy directly from high-dimensional observations, which is formulated as a partially observable Markov decision process (POMDP) with a tuple $(\mathcal S,\mathcal A,\mathcal O,\mathcal T,\mathcal R)$. Here, $\mathcal S$ is the state space, $\mathcal A$ is the action space, $\mathcal O$ is the observation space, $\mathcal R(s_t, a_t)$ is the reward function, and $ \mathcal T(s_{t+1} \mid s_t, a_t) $ is the state-transition distribution. 
In this setting, the agent cannot access the true states in $\mathcal S$.
At each timestep $t\in[1; T]$, the agent takes an action $a_t \in A$ to interact with the environment and receives a reward $r_t = \mathcal R(s_t, a_t)$. The objective is to learn a policy that maximizes the expected cumulative reward $\mathbb E_p[\sum^T_{\tau=1} r_{\tau}]$.

To improve policy learning and sample efficiency of visual RL, we aim to transfer previous knowledge from multiple offline tasks. The offline datasets are reward-free and exclusively consist of image-action pairs $\{(o_t,a_t)\}$. 
It is important to note that there might be substantial distribution shifts in observations ($\mathcal O$), state transition functions ($\mathcal T$), and behaviors ($\mathcal A$) across task domains, which pose significant challenges in transfer learning, providing strong motivation for the development of a dynamic domain-selective transfer RL approach. 
The primary goal of our approach is to efficiently bridge the gap between tasks in terms of state representations, physical dynamics, and action behaviors.

\section{Method}

In this section, we present a comprehensive overview of the pretraining process in the source datasets and the subsequent transfer learning process in the target task. 
The transfer learning process consists of two stages, \textit{i.e.}, \textit{domain-selective dynamics transfer} and \textit{behavior learning with generative action replay}, as shown in Fig. \ref{fig:method} and described in detail in \cref{algo:algorithm}.

\subsection{Why model-based RL for domain transfer?}
Our overall pipeline is built upon model-based RL, which involves learning the underlying dynamics from a buffer of past experiences, optimizing the control policy through future rollouts of compact model states, and executing actions in the environment to append the experience buffer. More precisely, we introduce a transfer RL approach based on the model-based DreamerV2 method~\cite{hafner2020mastering}. 
Unlike previous work, the world model in our approach serves not only as a simulator for policy learning but also provides a measure of task relevance for both dynamics representation transfer and behavior transfer discussed in the following sections. Additionally, after pretraining the source world model, subsequent algorithms can rely on the fixed parameters of this model, making it more universal in real-world scenarios and decoupled from the source data.

\subsection{Multi-Task Offline Pretraining}

\begin{figure}[t]
\begin{center}
\centerline{\includegraphics[width=\columnwidth]{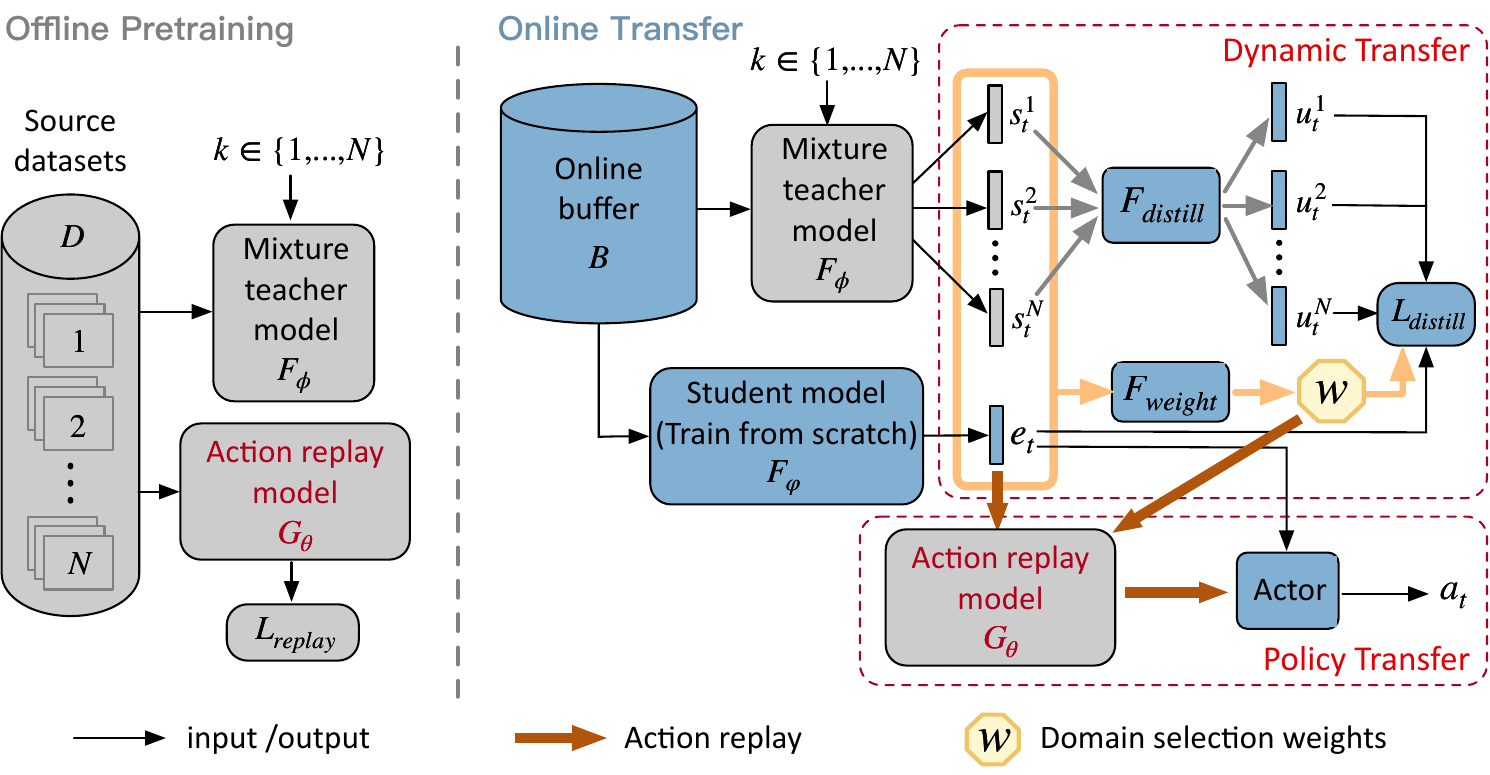}}
\caption{\textbf{Left:} We employ multiple offline domains to train a mixture world model ($F_\phi$), whose parameters are frozen during the subsequent transfer learning process. \textbf{Right:} In the target domain, we use $F_\phi$ as the teacher model and dynamically distill prior knowledge from it with a set of domain-similarity weights $\mathcal{W}$. These weights are further used to reproduce the most relevant source actions to guide the target policy.}
\label{fig:method}
\end{center}
\vspace{-20pt}
\end{figure}

\noindent \textbf{Mixture world model as the teacher model.} 
As illustrated in Fig. \ref{fig:method}, we consider multiple reward-free, action-conditioned datasets denoted as $\mathcal{D}$. These datasets comprise expert data that has been previously collected from $N$ tasks and is readily available for our use.
Initially, we pretrain an action-conditioned video prediction model, denoted as $F_{\phi}$, with the explicit task label $k\in\{1,\ldots,N\}$.
In contrast to APV~\cite{seo2022reinforcement}, an existing model-based pretraining-finetuning transfer RL method, our approach incorporates actions during the pretraining phase, which is reasonable in learning the consequences of state transitions.
The pretrained models consist of three main components as follows:
\begin{equation}
\label{eq:pre_train_model}
  \begin{split}
  \text{Representation model:} & \quad q(s_t \mid s_{t-1}, a_{t-1}^k, o_t^k, k) \\
  \text{Dynamics model:} & \quad p(\hat{s}_t \mid s_{t-1}, a_{t-1}^k, k) \\
  \text{Decoder model:} & \quad  p(\hat{o}_t \mid s_{t}, k). \\
  \end{split}
\end{equation}
The representation model extracts posterior latent states $s_t$ from observations $o_t$, previous states $s_{t-1}$, previous actions $a_{t-1}$ and task label $k$. The dynamics model follows the Recurrent State Space Model (RSSM) architecture from PlaNet~\cite{hafner2019learning} to predict the prior latent states $\hat{s}_t$ without access to the corresponding $o_t$. The decoder reconstructs $\hat{o}_t$ given the latent states. For task $T_k$,  all components are optimized jointly using the following loss function:
\begin{equation}
\label{eq:pre_train_loss}
\begin{split}
\mathcal{L}_\text{source} =  &\mathbb{E} \ \{
\sum_{t=1}^{T} \underbrace{-\ln p(\hat{o}_t \mid s_{t}, k)}_{\text {Image reconstruction}} \\
&+\underbrace{\beta \ \mathrm{KL}[q(s_{t} \mid s_{t-1}, a_{t-1}^k, o_{t}^k, k) \parallel p(\hat{s}_{t} \mid s_{t-1}, a_{t-1}^k, k)]}_{\text{KL divergence}}\},
\end{split}
\end{equation}
where $\beta$ is a hyperparameter of the Kullback-Leibler (KL) divergence that regularizes the approximate posterior learned from the representation model toward the prior learned from the dynamics model.

\vspace{5pt}
\noindent \textbf{Behavior replay.} 
We simultaneously utilize the offline source datasets to learn an action replay model to guide subsequent target behavior learning.
Inspired by BCQ~\cite{fujimoto2019off}, which is an offline RL method, we design $G_\theta$ using a state-conditioned variational auto-encoder (VAE)~\cite{kingma2013auto,sohn2015learning}. 
The action replay model $G_\theta$ consists of an encoder $E_{\theta1}$ and a decoder $D_{\theta2}$. 
The encoder takes a state-action pair and a task label $k$, and outputs a Gaussian distribution $\mathcal{N}(\mu, \sigma)$. The state $s$, along with a latent vector $z$ sampled from the Gaussian distribution and a task label $k$, is passed to the decoder $D_{\theta2}$ which outputs an action:
\begin{equation}
    \label{eq:VAE}
    \mu, \sigma = E_{\theta1}(s, a, k), \quad \hat{a}=D_{\theta2}(s,z,k), \quad z\sim \mathcal{N}(\mu, \sigma).
\end{equation}
The action replay model $G_\theta$ is optimized by
\begin{equation}
    \label{eq:VAE_loss} 
    \mathcal{L}_\text{replay} = \mathbb{E} \ \Big[ \sum_{(s,a)\in D} (a-\hat{a})^2 + \mathrm{KL}\left(\mathcal{N}(\mu, \sigma) \parallel \mathcal{N}(0, 1)\right)\Big].
\end{equation}

\begin{algorithm}[!t]
  \caption{\model{} with improved \textcolor{MyDarkRed}{dynamics learning,} \textcolor{MyDarkGreen}{behavior learning} \textcolor{MyDarkBlue}{\& policy deployment}
  }
  \label{algo:algorithm}
  \small
  \SetAlgoLined
  \textbf{Hyperparameters: }{$H$: Imagination horizon}
%   \KwResult{how to write algorithm with \LaTeXe }
\DontPrintSemicolon

  % Initialize the offline replay buffer $\mathcal{D}$ with multiple tasks.\\
  Initialize the online replay buffer $\mathcal{B}$ with random episodes. \\
  \While{not converged}{
    \For{update step $c = 1 \dots C$}{
        Draw data sequences $\left\{\left(o_{t}, a_{t}, r_{t}\right)\right\}_{t=1}^{T} \sim \mathcal{B}$. \\
        \texttt{// Dynamics learning} \\
        \textcolor{MyDarkRed}{Compute distillation loss using \cref{eq:distillation_loss} and update world model parameters using \cref{eq:all_loss}} \\
        % \textcolor{MyDarkRed}{Draw data sequences $\left\{\left(o_{t}, a_{t}\right)\right\}_{t=1}^{T} \sim \mathcal{D}$ from each source task.} \\
        % \textcolor{MyDarkRed}{Compute action generation model loss using \cref{eq:VAE_loss} and update model parameters.} \\
        % 
        \texttt{// Behavior learning} \\
        \For{time step $i = t \dots t+H$}{
        \textcolor{MyDarkGreen}{Select the task label $k$ with highest confidence in \cref{eq:weight} }\\
        \textcolor{MyDarkGreen}{Imagine an action ${a}_i \sim \pi (a_i \mid e_i, G_\theta(e_i, k))$} \\
        Predict rewards ${r}_i \sim p(\hat{r}_i \mid e_i)$ and values $v_\psi(e_i)$
        }
        Update the actor and value models in \cref{eq:ac-model} using estimated rewards and values.
    }
    \texttt{// Environment interaction} \\
    $o_{1} \leftarrow$ \texttt{env.reset}() \\%\tcc*[r]{Environment interaction}
    \For{time step $t = 1\dots T$}{
    Calculate the posterior state $e_{t} \sim q\left(e_{t} \mid e_{t-1}, a_{t-1}, o_{t}; \varphi \right)$ from history. \\
    \textcolor{MyDarkBlue}{Use the teacher model to obtain $\{\hat{s}_t^i \sim p(e_{t-1}, a_{t-1}, i; \phi) \mid i\in[1,N]\}$ and determine the task label $k$ with highest confidence in \cref{eq:weight} }\\
    \textcolor{MyDarkBlue}{Compute ${a}_t \sim \pi (a_t \mid e_t, G_\theta(e_t, k))$}\\
    $r_t, o_{t+1} \leftarrow$ \texttt{env.step}($a_t$)
    }
    Add experience to the online replay buffer $\mathcal{B} \leftarrow \mathcal{B} \cup\{\left(o_{t}, a_{t}, r_{t}\right)_{t=1}^{T}\}$.
  }
\end{algorithm}

\subsection{Domain-Selective Dynamics Transfer}

It is important to note that, even though the pretraining method is effective, a simple finetuning approach may encounter challenges due to the potential discrepancy in visual observations, physical dynamics, or even action spaces across task domains.
Therefore, when a novel target task emerges, we initialize a student world model $F_\varphi$ from scratch, while freezing the parameters of the teacher model $F_{\phi}$ to transfer the dynamics representations from the source domains (see Fig. \ref{fig:method}). 
In addition to the model components outlined in \cref{eq:pre_train_model}, $F_\varphi$ also incorporates a reward model represented as $\hat{r}_t \sim p(\hat{r}_t \mid s_t)$. 
To avoid confusion of notations, we use $s_t^k$ to denote the state obtained from the teacher model with task label $k$, and $e_t$ to denote the state of the target student model.
Given a latent state denoted by $e_{t-1}$ and a corresponding action $a_{t-1}$, we first transit this state to the next time step individually using the teacher model and the student model, obtaining $\{s_t^k \sim p(e_{t-1}, a_{t-1}; \phi)\}_{k=1}^N$ and $e_t \sim p(e_{t-1}, a_{t-1}; \varphi)$.

To close the distance between the marginal distributions of state transitions produced by the student world model and the dynamics estimated by the teacher model, we incorporate a distillation network in $F_\varphi$, denoted as $F_\text{distill}$, which takes the form of a multilayer perceptron (MLP). 
The role of this module is to extract transferable features from the predicted states of the teacher model. In other words, it transforms the states $s_t^k$ predicted by the teacher model into a set of transferable features $\{u_{t}^k = F_\text{distill}(s_t^k)\}_{k=1}^N$. These features are then used in the knowledge distillation loss.

Intuitively, each source task may hold varying impacts on the dynamics learning of the target visual control task. We introduce the concept of domain-similarity weights and propose to optimize these weights through the knowledge distillation loss. 
By learning this set of weights, we can dynamically transfer knowledge in an adaptive manner based on offline-online task relevance.
To compute the similarity weight $\mathcal{W}$, we concatenate the predicted state $s_t^k$ of teacher model  and the predicted state $e_t$ of the student model. 
This concatenated representation is then fed into a fully-connected layer $F_\text{weight}$, followed by a softmax activation function:
% \vspace{1pt}
\begin{equation}
    \label{eq:weight}
    \text{Domain selection:} \quad \mathcal{W} = \{w_k\}_{k=1}^N = \text{Softmax}(\{F_\text{weight}(s_t^k \ast e_t)\}_{k=1}^N),
\end{equation}
where $\ast$ denotes the operation of concatenation. 
In order to avoid the collapse of domain-specific weights, wherein $w_i=1$ when $i=c$ and $w_i=0$ for $i\neq c$, with $c$ denoting the offline task most akin to the present online task, we establish a minimum threshold of $0.1$ for the weights.
We then minimize the Euclidean distance between pairs of states as follows, taking into account the corresponding domain-similarity weights:
\begin{equation}
    \label{eq:distillation_loss}
    \mathcal{L}_\text{distill} = \sum_{k=1}^{N}  \sum_{t=1}^{T}  w_k \cdot \parallel e_t - u_t^k \parallel_2^2.
\end{equation}
The overall objective of the student model can be written as follows, where $\alpha$ is a hyperparameter: 
% \vspace{1pt}
\begin{equation}
\begin{split}
    \label{eq:all_loss}
    \mathcal{L}_\text{target} =  \mathbb{E} \ \Big[\Big[ &
\sum_{t=1}^{T} \underbrace{\beta \ \mathrm{KL}\big[q(e_{t} \mid e_{t-1}, a_{t-1}, o_{t}) \parallel p(\hat{e}_{t} \mid e_{t-1}, a_{t-1})\big]}_{\text {KL divergence}} \\
& \underbrace{-\ln p(\hat{o}_{t} \mid e_{t})}_{\text {Image reconstruction }} \underbrace{-\ln p(\hat{r}_{t} \mid e_{t})}_{\text {Reward prediction}} \Big] + \alpha \ \mathcal{L}_\text{distill}\Big].
\end{split}
\end{equation}

\cref{eq:distillation_loss} is the fundamental basis for \model{}.
When the dynamics of the source domain are similar to the target task, the latter term of this loss naturally becomes smaller. On the other hand, for source tasks with significantly different dynamics from the target task, the model will minimize the weight term to minimize this loss.
The \textit{domain-selective} distillation loss enables the student model to adaptively learn from the teacher model, acquiring significant prior knowledge regarding intricate physical dynamics from the most relevant source tasks. By selectively distilling knowledge from these source tasks, the student model can adapt and incorporate valuable information to enhance its overall learning capabilities.

\subsection{Domain-Selective Behavior Transfer}

We utilize an actor-critic algorithm to learn the policy over the predicted future state and reward trajectories.
As shown in Fig. \ref{fig:method}, we use the action replay model $G_\theta$ to promote policy learning, which \textit{1) provides an efficient indication when a strong correlation exists between the source and target tasks}, and \textit{2) expends exploration of action space when there is little correlation between them.}
The parameters in $G_\theta$ are frozen at this stage. 
Reusing the similarity weights learned in dynamics transfer, we can dynamically select task label $k$ with the highest confidence to generate action guidance. 
We exclusively employ the decoder $D_{\theta2}$ of action generation model $G_\theta$ to replay source-domain actions, which takes the state $e_t$ of the student model and the selected task label $k$ with highest confidence as inputs.
We modify the actor model and the value model as follows:
\begin{equation}
\begin{split}
\label{eq:ac-model}
    &\text{Actor model:} \quad {a}_{t} \sim \pi(a_{t} \mid e_t, \textcolor{blue}{G_\theta(e_t, k)}), \\
    &\text{Value model:} \quad v_{\psi}(e_{t}) \approx \mathbb{E}_{\pi\left(\cdot \mid e_{t}, G_\theta(e_t, k)\right)} \sum_{t^{\prime}=t}^{t+H} \gamma^{t^{\prime}-t} r_{k},
\end{split}
\end{equation}
where $H$ is the imagination time horizon and $\gamma$ is the reward discount. The actor model is optimized to maximize the value estimation, while the value model is optimized to approximate the expected imagined rewards. 
The training target for the value model is:
\begin{eqnarray}
    V_{t} = r_{t} + \gamma
    \begin{cases}
        (1 - \lambda) v_{\psi}(e_{t+1}) + \lambda V_{t+1} & \text{if} \quad t < H, \\
         v_{\psi}(e_{H}) & \text{if} \quad t = H,
    \end{cases}
\end{eqnarray}
where $\lambda$ equals to $0.95$. 
Similar to the process of behavior learning, we also utilize the action replay model $G_\theta$ to draw action from the actor model during policy deployment. As shown in Lines 20-21 in \cref{algo:algorithm}, the action guidance is dependent on current states $e_i$ and the source task label with the highest domain-similarity weights, which may evolve over time.

\section{Experiments}

\subsection{Experimental Setup}

\begin{figure}[t]
\begin{center}
\centerline{\includegraphics[width=0.9\columnwidth]{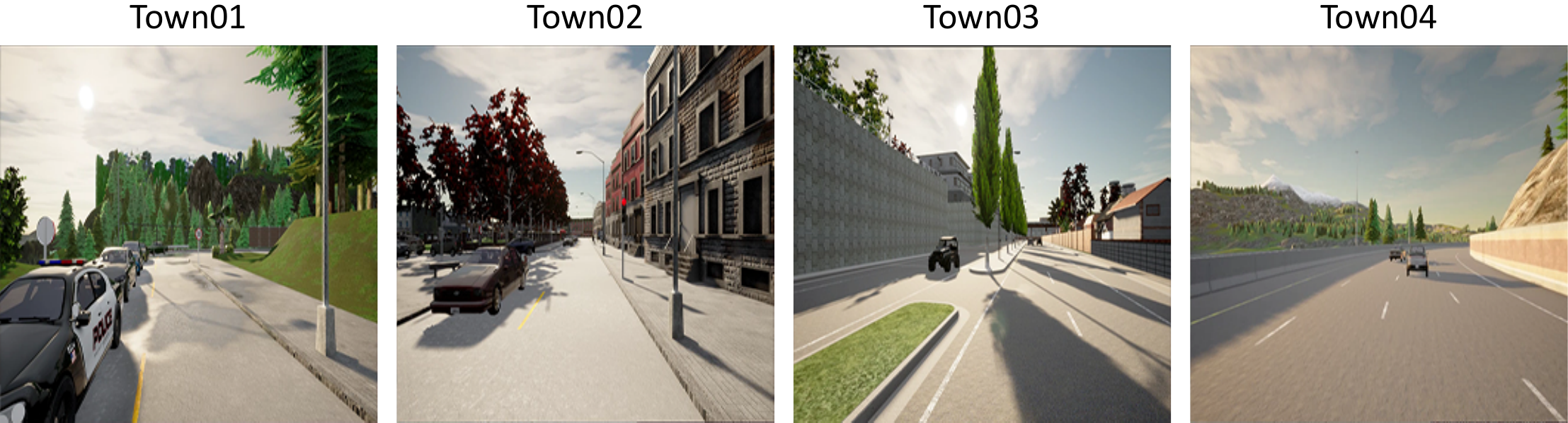}}
\caption{Showcases of selected towns in CARLA environment.}
\label{fig:showcases_carla}
\end{center}
\vspace{-20pt}
\end{figure}

\noindent \textbf{Benchmarks.} We evaluate \model{} on three visual RL environments in an offline-to-online domain transfer setup:
\begin{itemize}[leftmargin=*]
    \item
    \textbf{Meta-World}~\cite{yu2020meta}: It simulates $50$ manipulation tasks,  all involving the same robotic arm. We collect $6$ offline datasets using expert experiences from \textit{button press topdown}, \textit{door open}, \textit{drawer close}, \textit{peg insert side}, \textit{pick place}, and \textit{push}. Each of them contains $10$ demonstrations. 
    \vspace{3pt}\item
    \textbf{DeepMind Control Suite} \cite{tassa2018deepmind}: It is a standard benchmark for visual-based RL that contains a diverse set of continuous control tasks.
    We collect offline datasets from $4$ tasks, \textit{i.e.}, \textit{cheetah run}, \textit{hopper stand}, \textit{walker walk}, and \textit{walker run}.  
    Each task contains $50$ trajectories of expert experiences.
    \vspace{3pt}\item 
    \textbf{CARLA} \cite{DBLP:conf/corl/DosovitskiyRCLK17}: It is an open-source simulator that provides more intricate and lifelike visual observations for research in autonomous driving. The objective of the agent is to maximize its driving distance within 1000 time steps while avoiding collisions with 30 other moving vehicles or barriers. As a result, the episode length is 1000 steps with the action repeat of 4. To encourage highway progression and penalise collisions, the reward is formulated as: $r_t=v^T_{ego}\hat{u}_h \cdot \Delta t -\xi_1 \cdot \mathbb{I} - \xi_2 \cdot |steer| $, where $v_{ego}$ represents the velocity vector of the ego-vehicle, projected onto the highway’s unit vector $\hat{u}_h$, and multiplied by time discretization $\Delta t = 0.05$ to measure highway progression in meters. The impulse $\mathbb{I} \in \mathbb{R^+}$ indicates the impact caused by collisions, and a steering penalty $steer \in [-1,1]$ aids in maintaining lane position. The visualization samples of four towns utilized in our experiments are shown in Fig. \ref{fig:showcases_carla}.
\end{itemize}

\vspace{5pt}
\noindent \textbf{Compared methods.}
% We compare our approach with five baselines, \textit{i.e.}, DreamerV2 \cite{hafner2020mastering}, APV \cite{seo2022reinforcement}, Iso-Dream \cite{pan2022iso}, SMART \cite{sun2023smart}, and TD-MPC2 \cite{hansen2023td}. 
%
We compare \model{} with the following approaches:
\begin{itemize}[leftmargin=*]
    \item \textbf{DreamerV2}~\cite{hafner2020mastering}: A model-based RL method that learns the policy directly from latent states in the world model. The latent representation enables agents to imagine thousands of trajectories simultaneously.
    \vspace{3pt}\item \textbf{APV}~\cite{seo2022reinforcement}: A model-based RL method that stacks an action-conditional RSSM model on top of the pretrained action-free RSSM model. We train this model by following its two-step training setting. 
    \vspace{3pt}\item \textbf{Iso-Dream}~\cite{pan2022iso}: A strong baseline for visual RL that learns different dynamics based on controllability. It rolls out noncontrollable states into the future and performs policy optimization based on the decoupled latent imaginations.
    \vspace{3pt}\item \textbf{SMART}~\cite{sun2023smart}: A generic multi-task pretraining framework that designs a Control Transformer coupled with a control-centric pretraining objective in a self-supervised manner.
    \vspace{3pt}\item \textbf{TD-MPC2}~\cite{hansen2023td}: A model-based RL method that primarily uses state information to learn task-oriented latent dynamics model purely from rewards, ignoring nuances unnecessary for the task at hand.
\end{itemize}

It is reasonable to compare our method with APV and Iso-Dream, as they are also built upon DreamerV2. Furthermore, our proposed transfer RL techniques can also be seamlessly integrated with DreamerV3 \cite{hafner2023mastering}, enhancing its overall performance. In this paper, our method is based on DreamerV2 unless otherwise specified.

% \vspace{5pt}
% \noindent More details about the benchmarks and baselines can be found in the supplementary materials.

\begin{figure}[!t]
\begin{center}
\centerline{\includegraphics[width=0.9\linewidth]{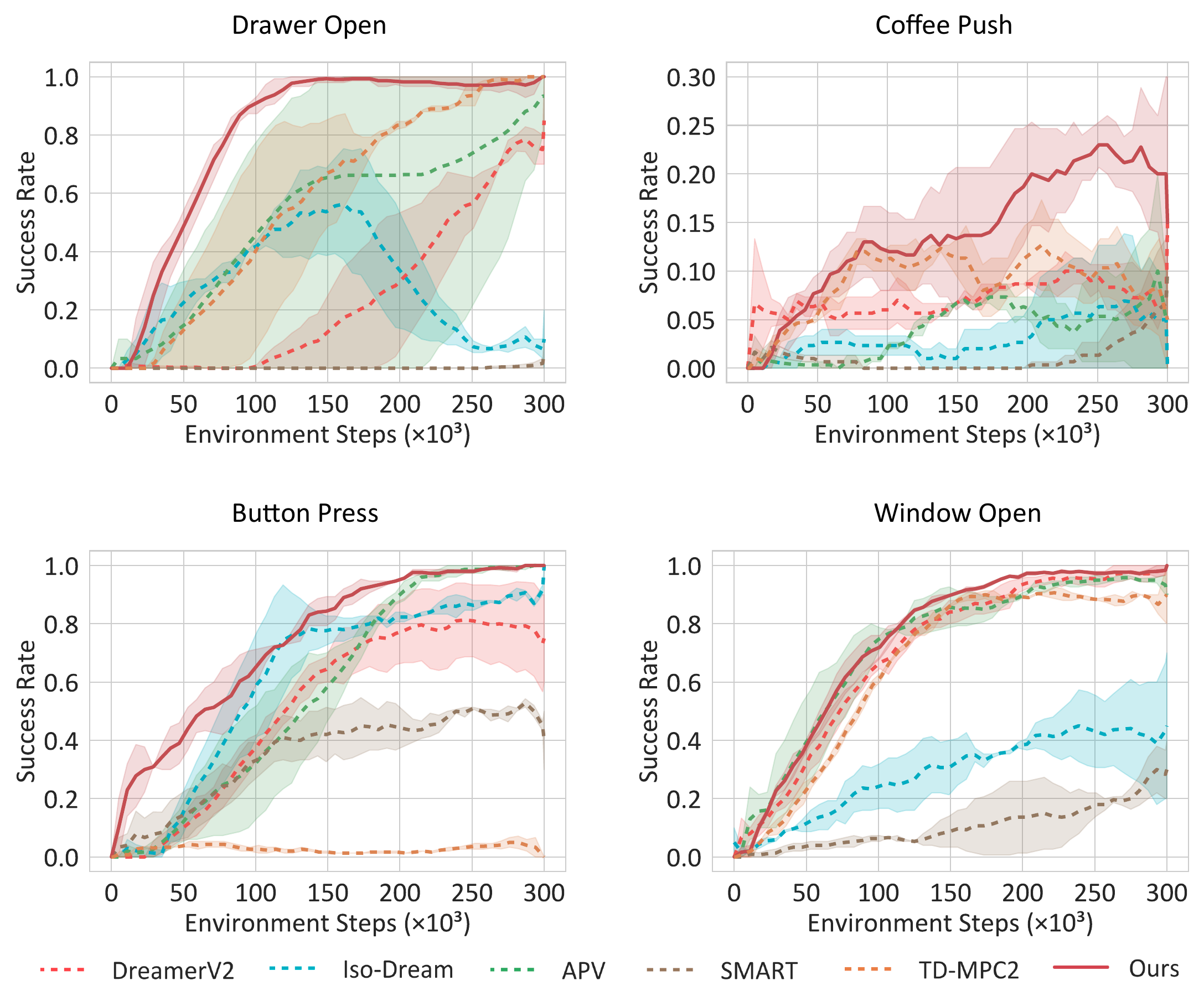}}
\caption{Performance comparison with the state-of-the-art methods on Meta-World as measured on the success rate. \model{} outperforms the compared models.}
\label{fig:meta_main}
\end{center}
\vspace{-20pt}
\end{figure}

\subsection{Main Results}

\noindent \textbf{Meta-World.}
We first pretrain the teacher model of the action-conditioned video prediction model by minimizing the objective in \cref{eq:pre_train_loss} for $200K$ gradient steps. 
The hyperparameters $\beta$ and $\alpha$ are set to $1$ in \cref{eq:all_loss}.
Our model is evaluated in $4$ tasks, \textit{i.e.}, \textit{drawer open}, \textit{coffee push}, \textit{button press}, and \textit{window open}. In all tasks, the episode length is $500$ steps without any action repeat. The number of environment steps is limited to $300K$. We run all tasks with $3$ seeds and report the mean success rate and standard deviations of $10$ episodes.
As shown in Fig. \ref{fig:meta_main}, our \model{} generally outperforms other methods on four tasks. 
Specifically, we improve DreamerV2 $20\%$ in \textit{drawer open} and $90\%$ in \textit{coffee push}.
TD-MPC2, despite its ability to handle state information effectively, exhibits weaker performance than our model when processing visual image inputs.

\begin{figure}[!t]
\begin{center}
\centerline{\includegraphics[width=0.9\columnwidth]{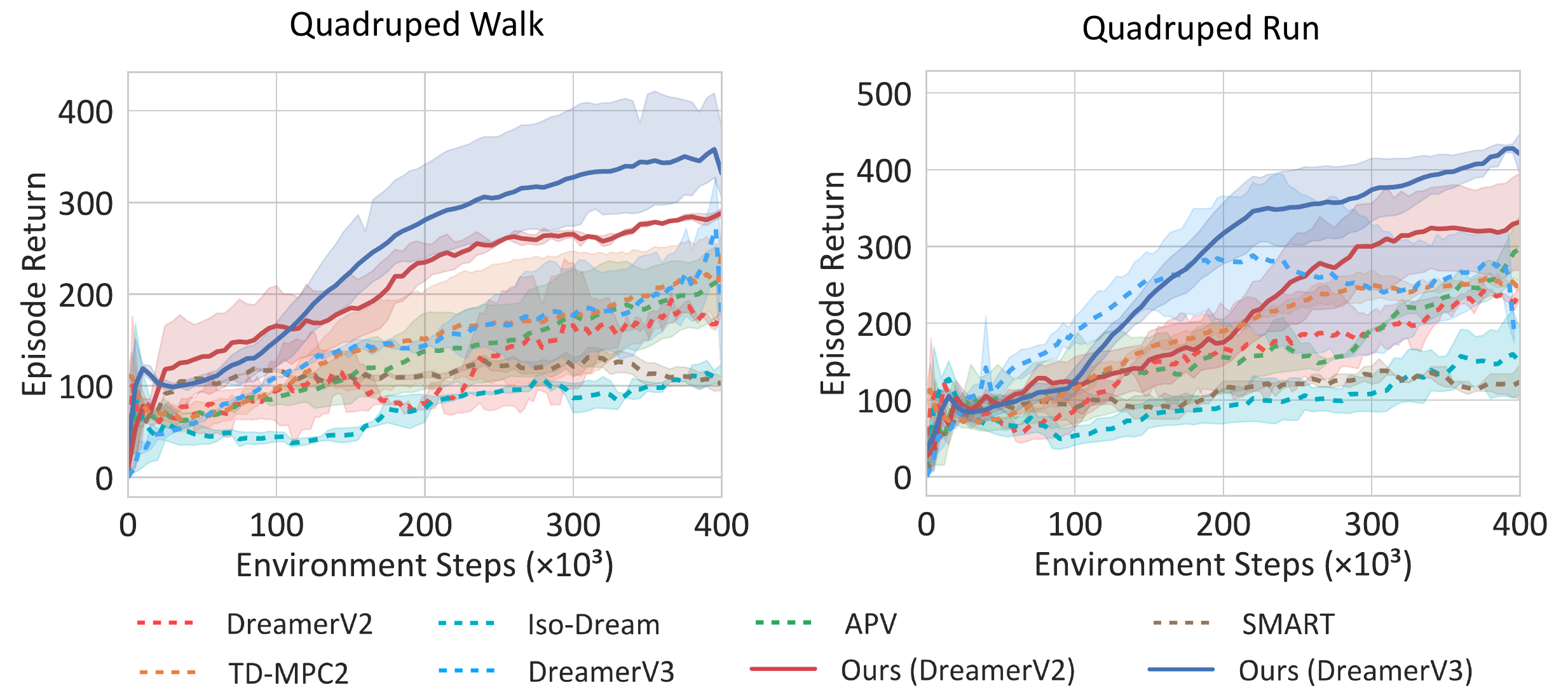}}
\caption{Performance comparison on two tasks from DeepMind Control Suite as measured on the episode rewards. Our \model{} with dynamic knowledge distillation achieves significant improvements compared with existing model-based RL approaches.}
\label{fig:compare_sota_dmc}
\end{center}
\vspace{-20pt}
\end{figure}

\vspace{5pt}
\noindent \textbf{DeepMind Control Suite.}
In this environment, the episode length is $1{,}000$ steps with the action repeat of $2$, and the reward ranges from $0$ to $1$.
For the online target tasks, we train our method for $200K$ iterations, which results in $400K$ environment steps. 
We evaluate \model{} with baselines on the mean episode rewards and standard deviations. The results of \textit{quadruped walk} and \textit{quadruped run} are illustrated in Fig. \ref{fig:compare_sota_dmc}.
Our framework achieves significant improvements compared with existing model-based RL approaches. For example, \model{} performs nearly $100$ higher performance on the task of \textit{quadruped walk} and \textit{quadruped run} than DreamerV2 after $400k$ steps environment interactions. 
Iso-Dream, which serves as a robust baseline for addressing visual control tasks through isolated state transition branches, exhibits limitations in handling these two tasks.
Compared with APV, which only uses the pretrained action-free world model as initialization to train downstream tasks, \model{} is encouraged to learn more precise state transitions based on action input and more useful source dynamics based on domain-selective knowledge distillation. Moreover, the learned domain selection weights help the agent adaptively transfer potentially useful action demonstrations from offline datasets. 
In addition, we utilize DreamerV3 as the network backbone and observe that our proposed techniques can be seamlessly integrated with DreamerV2/V3 and consistently enhance their performance.

\subsection{Ablation Studies}
\begin{figure}[!t]
\begin{center}
\centerline{\includegraphics[width=0.9\linewidth]{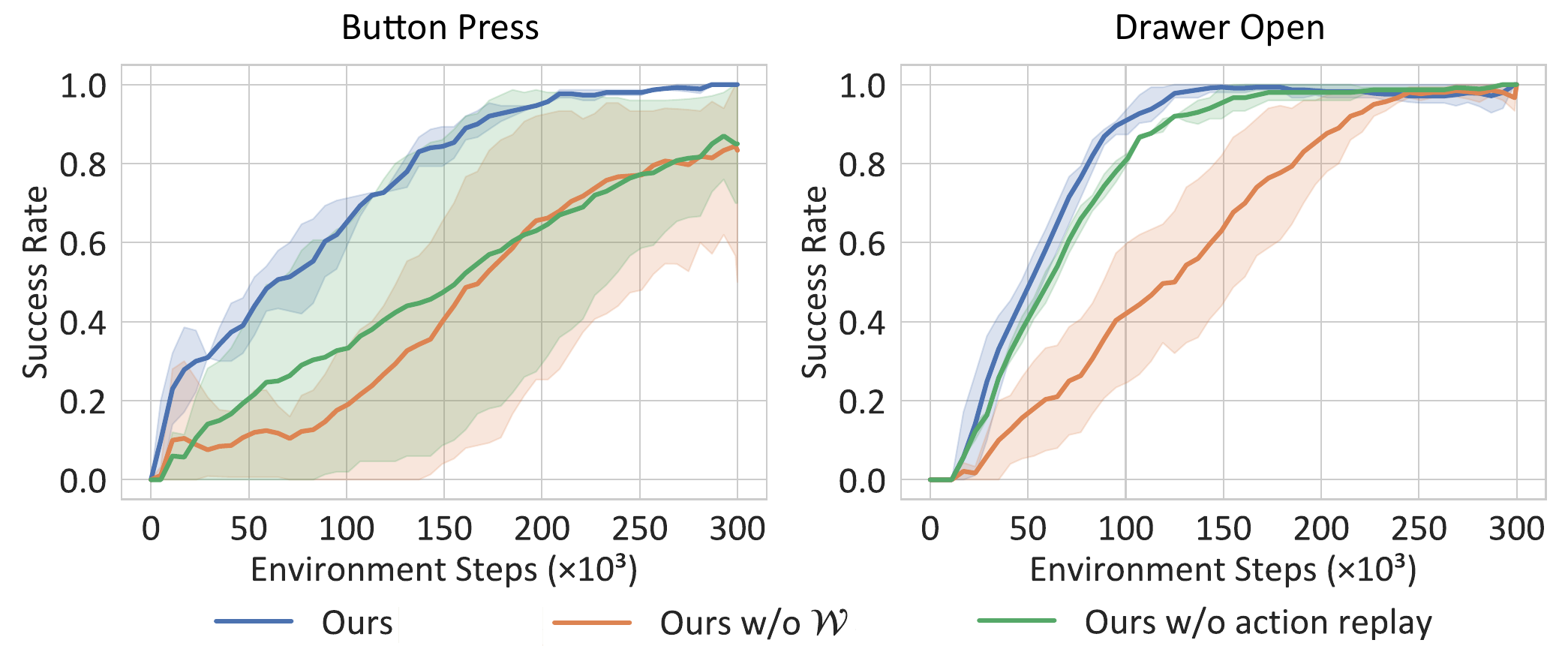}}
\caption{Ablations of \model{} that illustrate the impact of learning time-varying domain selection weights and optimizing behavior learning with action replay.}
\label{fig:ablation_meta}
\end{center}
\vspace{-10pt}
\end{figure}

We conduct ablation studies to confirm the validity of learning a set of time-varying domain selection weights and behavior learning with action replay on two tasks, as shown in Fig. \ref{fig:ablation_meta}.
Without the process of learning the importance weights (\textcolor{orange}{orange}) to measure the similarity between source and target tasks, the performance of our model has decreased by about $25\%$ in \textit{button press}, and it requires more timesteps to improve the behavior policy in \textit{drawer open}.
It demonstrates that information in different source tasks has different impacts on the target task, and a domain-selective knowledge distillation loss with importance weights encourages the student model to adaptively find useful prior knowledge and transfer it to help the dynamics learning in downstream tasks.
Moreover, we evaluate \model{} without action replay model for behavior learning (\textcolor{MyDarkGreen}{green}). The result shows that our proposed domain-selective behavior learning strategy can identify potentially valuable source actions and employ them as exemplar guidance for the target policy. 

\subsection{Analyses of Task Relations}

\begin{figure}[!t]
\begin{center}
\centerline{\includegraphics[width=0.9\linewidth]{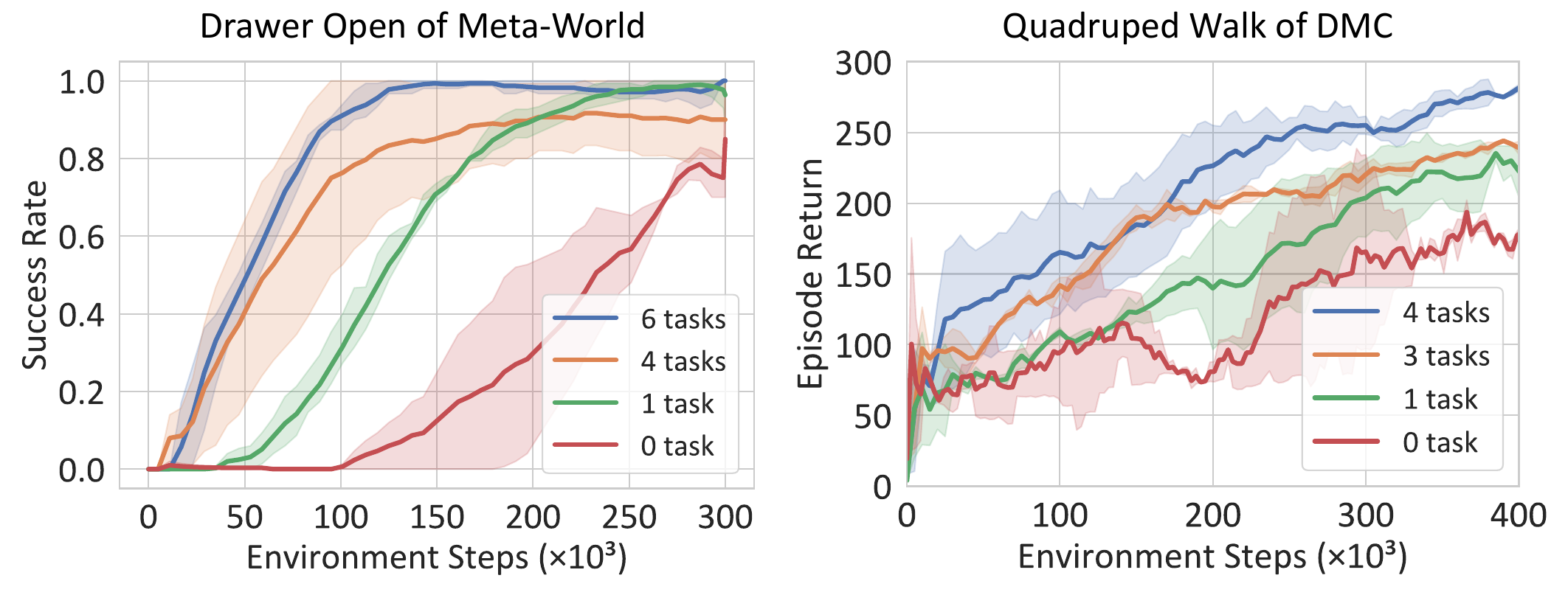}}
\caption{Analyses on the impact of different source task configurations. Compared with DreamerV2 (\textcolor{red}{$0$ source task}) which is exclusively trained on the target task, our method consistently achieves positive offline-to-online transfer even when it has access to only one task with the lowest importance weights (\textcolor{MyDarkGreen}{$1$ task}).}
\label{fig:ablation_number}
\end{center}
\vspace{-20pt}
\end{figure}

\begin{figure}[!t]
\begin{center}
\centerline{\includegraphics[width=0.9\linewidth]{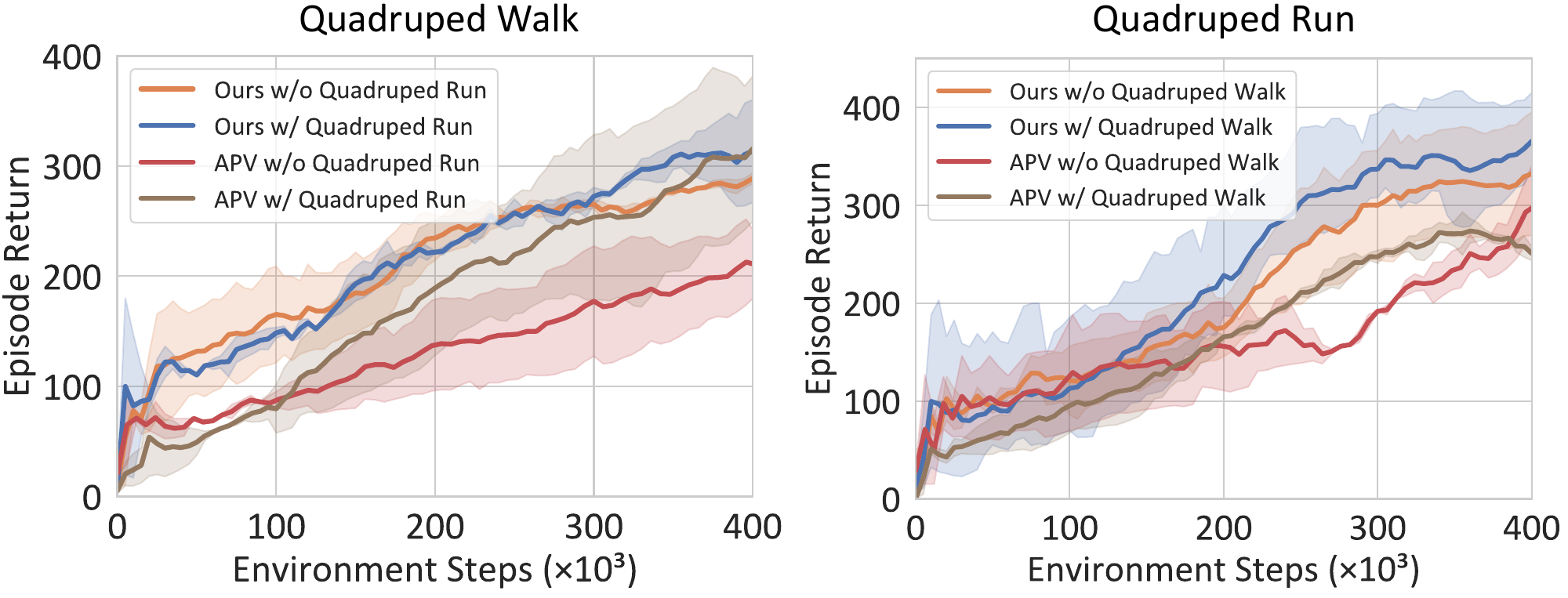}}
\caption{Analyses on the impact of same dynamics between offline source tasks and target task. Our model shows more stable performance compared with APV, eliminating the reliance on similar physical dynamics across domains.}
\label{fig:with_same_embody}
\end{center}
\vspace{-20pt}
\end{figure}

\noindent \textbf{Impact of fewer or less-relevant source tasks.}
To analyze the robustness of our approach to different source domain configurations, we sequentially decrease the number of source domain tasks according to the learned importance weights, \textit{i.e.}, gradually removing the task with the highest importance weight. The results are shown in Fig. \ref{fig:ablation_number}. 
We have two observations in this figure. First, compared with the baseline model that is solely trained on the target task, our approach consistently achieves positive offline-to-online transfer even when it can only access parts of the source datasets with lower importance weights.
Second, as the number of the source tasks grows, the performance of \model{} improves as well, demonstrating its effectiveness in identifying task similarity and improving the target policy with the expanded offline datasets.

\vspace{5pt}
\noindent \textbf{Impact of various dynamics between source/target tasks.}
% Impact of the dynamics of target task in the source tasks
%
Furthermore, we use a setup where the underlying dynamics of the target task are already seen in the source domain, but the task is different from the source tasks, \textit{i.e.}, the reward functions are different.
Specifically, we add the task of \textit{quadruped walk} (\textit{quadruped run}) to the offline dataset and then transfer the knowledge to the task of \textit{quadruped run} (\textit{quadruped walk}). 
In Fig. \ref{fig:with_same_embody}, our model shows superior performance, regardless of the presence of similar dynamics between the source and target domains.  In contrast, APV is unstable and depends heavily on the similarity of physical dynamics across domains, such as \textit{quadruped walk}.

\begin{figure}[t]
\begin{center}
\centerline{\includegraphics[width=0.7\linewidth]{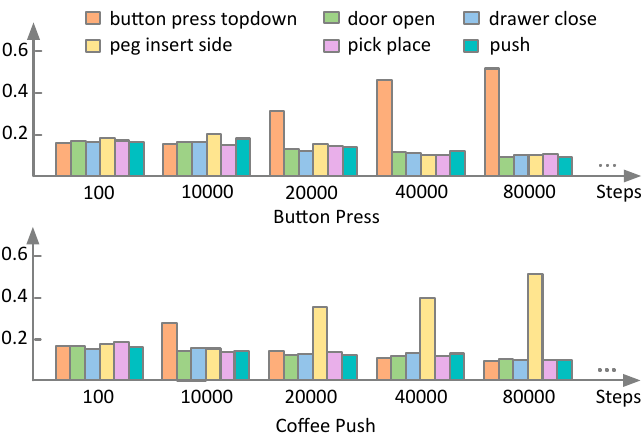}}
\caption{Weight distribution of different source tasks during the training phase. 
%Notably, our approach can discover the most relevant source task when it is converged.
}
\label{fig:weight}
\end{center}
\vspace{-10pt}
\end{figure}

\vspace{5pt}
\noindent \textbf{Changes in domain selection weights.} In Fig. \ref{fig:weight}, we show the weights of different source tasks during the training phase. For example, in the online \textit{button press} task, as the training progresses, the weight of \textit{button press todown} in source tasks increases and then becomes dominant. This shows that our model can dynamically transfer knowledge in an adaptive manner.

\begin{figure}[!t]
\begin{center}
\centerline{\includegraphics[width=0.55\columnwidth]{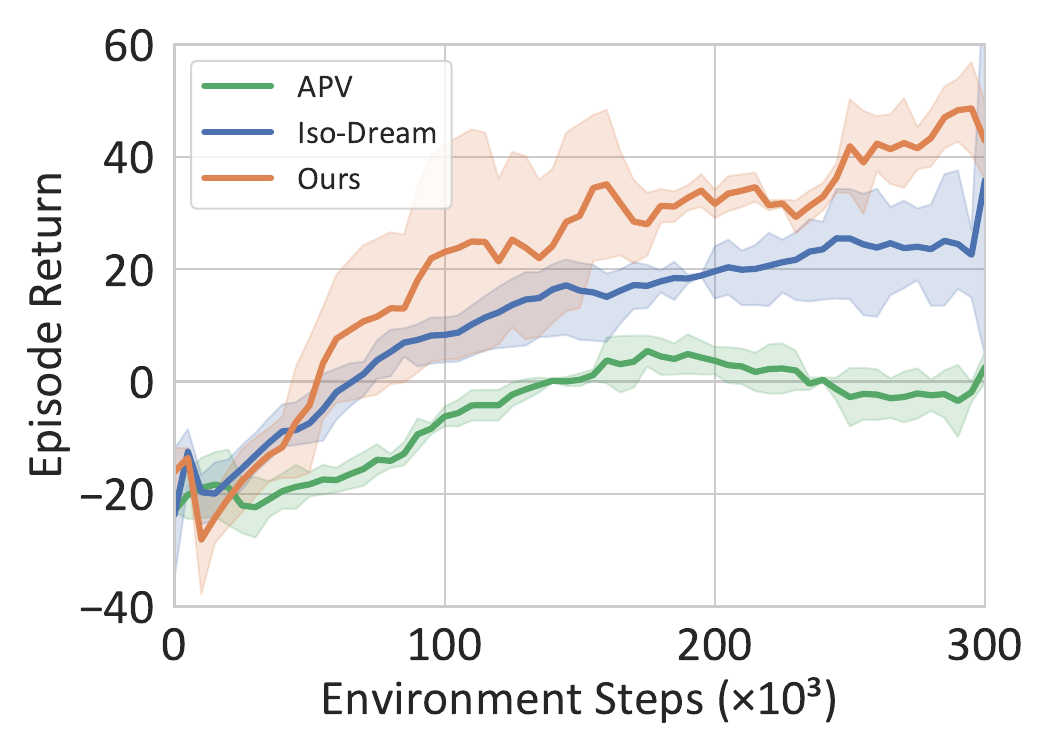}}
\caption{Performance comparison in CARLA environment as measured on the episode rewards. Our \model{} can improve the performance of Iso-Dream.}
\label{fig:carla}
\end{center}
\vspace{-20pt}
\end{figure}

\subsection{Results on CARLA environment}
We also demonstrate the performance of \model{} in CARLA. 
In our experiments, we use the expert datasets collected from three distinct maps, \textit{i.e.}, ``Town01'', ``Town02'', and ``Town03'', and evaluate our model in a first-person highway driving task in ``Town04''. The visualization samples can be found in the supplementary materials.
We employ Iso-Dream, a model demonstrated effective in the CARLA environment, instead of Dreamerv2 as our network backbone. Our method is trained for $75K$ iterations, resulting in $300K$ environment steps. The results are shown in Fig. \ref{fig:carla}. Compared with APV, which also uses offline datasets to pretrain, our model presents a remarkable advantage. Comparing the orange and blue curves, we see that our framework can improve the performance of Iso-Dream.

\subsection{Results with Medium Offline Data}

In this section, we further use the ``medium'' datasets as the source domains to verify the generalization and robustness of our \model. 
The medium datasets are generated by first training a policy online using DreamerV2, early-stopping the training, and collecting $200$ episodes from this partially-trained policy for each task.
We collect $6$ offline datasets in \textit{door close}, \textit{faucet open}, \textit{handle press}, \textit{plate slide}, \textit{reach wall}, and \textit{window close}.  
Our model is compared with the methods that also use offline datasets for pretraining the model, and the results are shown in Table \ref{tab:medium}. 
We can observe that our \model{} presents a remarkable advantage against other methods in terms of both success rate and episode return. 
In \textit{coffee push}, \model{} improves DreamerV2 by around $\textbf{75\%}$ ($0.20\rightarrow 0.35$) in success rate and by over $\textbf{70\%}$ ($328\rightarrow 561$) in episode return.
The results demonstrate that our model is not sensitive to the quality of source domain data, as it can achieve impressive performance even with medium datasets.

\begin{table*}[t]
  \centering
  \caption{Performance comparison, measured by the success rate and episode return, with baselines on the Meta-World environment using ``medium'' datasets as source domains.}
  \label{tab:medium}
    \setlength\tabcolsep{5pt}
    \begin{center}
    \begin{small}
    % \begin{sc}
    \begin{tabular}{l|ccccc}
    \toprule
    Methods  & DreamerV2 & APV & SMART  & Ours \\
    \midrule
    \multicolumn{5}{c}{Success Rate} \\
    \midrule
Coffee Push  &     0.20 $\pm$ 0.19      &  0.18 $\pm$ 0.15      &     0 $\pm$ 0        & \textbf{0.35 $\pm$ 0.15} \\
Drawer Open  &    0.62 $\pm$ 0.44       &   0.63 $\pm$ 0.37  &       0 $\pm$ 0      & \textbf{0.95 $\pm$ 0.08}  \\
    \midrule
    \multicolumn{5}{c}{Episode Return} \\
    \midrule
Coffee Push  &     328 $\pm$ 370      &   253 $\pm$ 789  &    14 $\pm$ 7      &  \textbf{561 $\pm$ 261} \\
Drawer Open  &     3726 $\pm$ 715      &   3942 $\pm$ 762   &       664 $\pm$ 116       &  \textbf{4276 $\pm$ 501}\\
    \bottomrule
    \end{tabular}
    % \end{sc}
\end{small}
\end{center}
\end{table*}

\section{Conclusion}

In this paper, we proposed a new domain-selective transfer learning framework called \model{} that improves visual RL with offline datasets with multiple tasks.
\model{} has two contributions. First, it provides a novel model-based pretraining and transfer learning pipeline for visual RL. Unlike APV~\cite{seo2022reinforcement}, it transfers action-conditioned dynamics from multiple source tasks with a set of importance weights learned by the world models.
Second, it provides a novel domain-selective behavior learning strategy that identifies potentially valuable source actions and employs them as exemplar guidance for the target policy.
Experiments in the Meta-World, DeepMind Control and CARLA environments demonstrated that \model{} significantly outperforms existing visual RL approaches.

Our \model{} has a limitation in the time consumption during offline pretraining, as the utilization of a mixture world model to simultaneously learn the dynamics of multiple tasks requires a significant amount of time for the model to converge.

\begin{credits}
% \subsubsection{\ackname} A bold run-in heading in small font size at the end of the paper is
% used for general acknowledgments, for example: This study was funded

\subsubsection{\discintname}
% It is now necessary to declare any competing interests or to specifically
% state that the authors have no competing interests. Please place the
% statement with a bold run-in heading in small font size beneath the
% (optional) acknowledgments,
% for example: The authors have no competing interests to declare that are
% relevant to the content of this article. Or: Author A has received research
% grants from Company W. Author B has received a speaker honorarium from
% Company X and owns stock in Company Y. Author C is a member of committee Z.
The authors have no competing interests to declare that are relevant to the content of this article.
\end{credits}

\section*{Acknowledgment}

This work was supported by the National Natural Science Foundation of China (Grant No. 62250062, 62106144), the Shanghai Municipal Science and Technology Major Project (Grant No. 2021SHZDZX0102), the Fundamental Research Funds for the Central Universities, and the CCF-Tencent Rhino-Bird Open Research Fund. 

\end{document}